\documentclass[10pt,twocolumn,letterpaper]{article}
\pdfoutput=1

\usepackage{cvpr}              

\usepackage[dvipsnames]{xcolor}
\usepackage{tabularx}
\usepackage{multirow}

\definecolor{cvprblue}{rgb}{0.21,0.49,0.74}
\usepackage[pagebackref,breaklinks,colorlinks,citecolor=cvprblue]{hyperref}

\def\paperID{19} 
\def\confName{CVPR}
\def\confYear{2024}

\title{Latent Directions: A Simple Pathway to Bias Mitigation in Generative AI}

\author{Carolina López Olmos$^{1,2}$ \quad 
Alexandros Neophytou$^{2}$ \quad 
Sunando Sengupta$^{2}$ \quad 
Dim P. Papadopoulos$^{1}$ \quad \\
$^1$ Technical University of Denmark \quad $^2$ Microsoft \\
{\tt\small  \{clopezolmos, alexandros.neophytou, sunando.sengupta\}@microsoft.com, dimp@dtu.dk
} }

\begin{document}
\maketitle
\begin{abstract}
Mitigating biases in generative AI and, particularly in text-to-image models, is of high importance given their growing implications in society. The biased datasets used for training pose challenges in ensuring the responsible development of these models, and mitigation through hard prompting or embedding alteration, are the most common present solutions. Our work introduces a novel approach to achieve diverse and inclusive synthetic images by learning a direction in the latent space and solely modifying the initial Gaussian noise provided for the diffusion process. Maintaining a neutral prompt and untouched embeddings, this approach successfully adapts to diverse debiasing scenarios, such as geographical biases. Moreover, our work proves it is possible to linearly combine these learned latent directions to introduce new mitigations, and if desired, integrate it with text embedding adjustments. Furthermore, text-to-image models lack transparency for assessing bias in outputs, unless visually inspected. Thus, we provide a tool to empower developers to select their desired concepts to mitigate. The project page with code is available online\footnote{\url{https://latent-debiasing-directions.compute.dtu.dk/}}.



\end{abstract}    
\section{Introduction}
\label{sec:intro}
Text-to-image models have enabled the possibility of generating personalized images with the content described by words, transforming industries, and even, our thoughts. Given these models’ impact on our lives, it is key to guarantee they are developed responsibly, battling stereotypes \cite{bloomberg, SolbesCanales2020SocializationOG}, lack of diversity, and inherited biases, but remaining truthful \cite{verge2024gemini}. Moreover, if this biased generated data is used for training future models, these biases will persist or be amplified in the new models themselves.

\begin{figure}[t]
  \centering
   \includegraphics[width=1\linewidth]{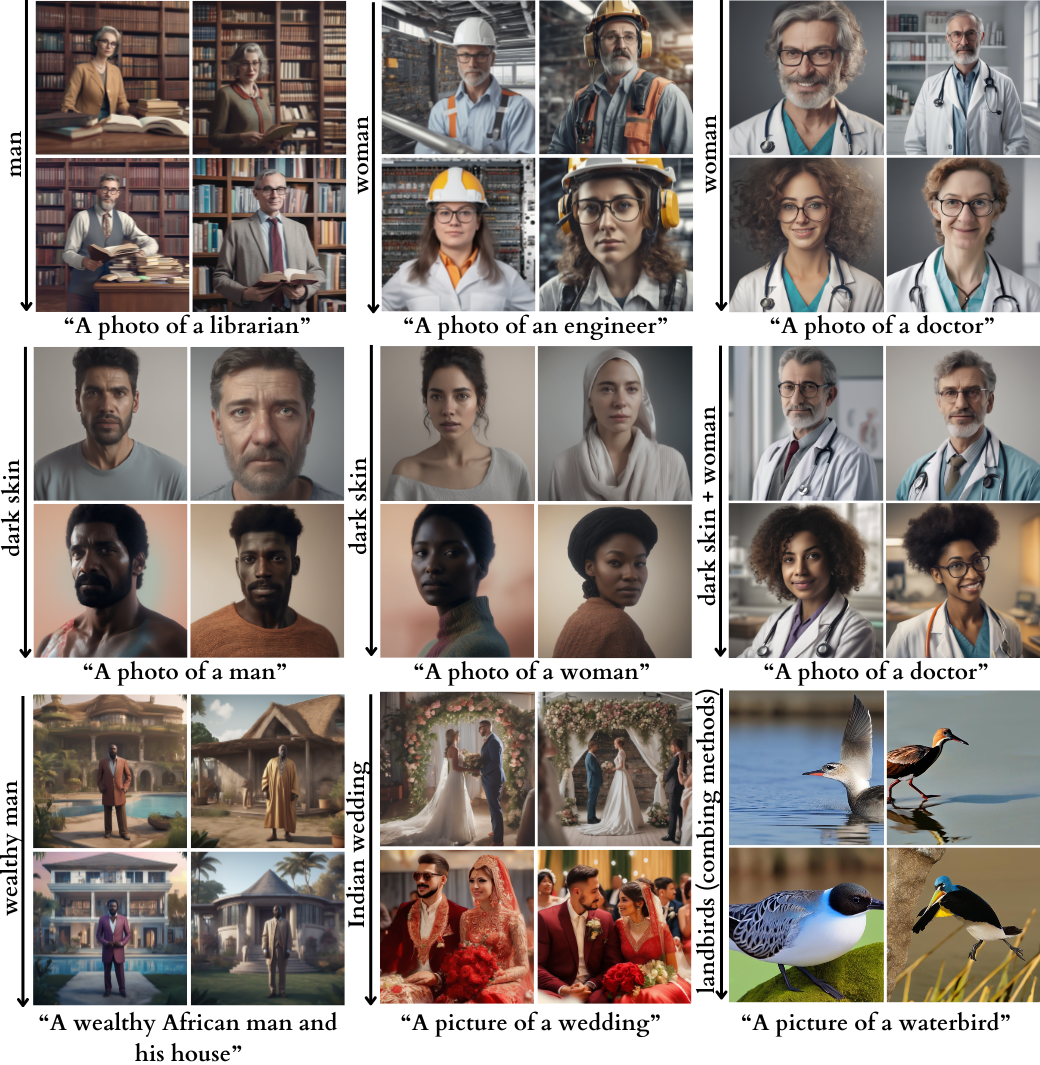}
   \caption{\textbf{Debiasing diverse concepts.} Generations observed upon the application of our approach in several scenarios.  }
   \label{fig:debiasing_results}
\end{figure}


From gender to race, to poor geographic representation, biases can be found everywhere in generative models. Leonardo N. and Dina B. \cite{bloomberg} have found a concerning pattern in Stable Diffusion v1.5 where low-paying jobs are dominated by women and darker-skinned individuals. While most of the work focuses on social and racial biases \cite{bolukbasi2016man, xu2018fairgan}, Basu \textit{et al.} \cite{basu2023inspecting} conduct a user study validating geographical biases in models like DALLE \cite{ramesh2021zeroshot} and Stable Diffusion \cite{rombach2022highresolution}, finding underrepresented 25 out of 27 countries. Cultural biases are also found when using homoglyphs in text-to-image synthesis \cite{mikolov2013exploiting}. Efforts to understand biases in vision-language models lead to the development of automated tools \cite{cho2023dalleval, luccioni2023stable}, the use of gender estimation \cite{li2023blip2}, face and skin tone detection \cite{Bulat_2017, feng2022racially}, and the evaluation of geographical representativeness using CLIP-based similarity and k-nearest neighbor models \cite{basu2023inspecting}. Mitigation efforts include prompt interventions \cite{bansal2022texttoimage}, the development of more inclusive datasets featuring images reflecting diverse geographic and socioeconomic contexts \cite{NEURIPS2022_5474d9d4, ramaswamy2023geode}, and alterations in prompt embedding strategies \cite{chuang2023debiasing, zhang2023itigen}. Chuang \textit{et al.} \cite{chuang2023debiasing} propose a method to mitigate bias by maximizing the similarity between biased and non-biased prompts. They construct a projection matrix to eliminate biased directions from text embeddings before inputting them into the model. Using a similar approach, but employing tokens of available image datasets, ITI-Gen \cite{zhang2023itigen} learns a set of prompt embeddings to append to the initial prompt.

We first present a tool to enhance developers’ visibility, given that we believe understanding the relationship between concepts, and the reason for certain attributes appearing in generations, is key to mitigating them. Secondly, we propose a straightforward novel approach for bias mitigation, linearly separating the latents, noisy information tensors of two different prompts, learning the transformation for debiasing in the latent space of the diffusion model. 

We apply this transformation, named \textit{latent direction}, at a specific weight, linearly combining it with the initial Gaussian noise. The results in a series of diverse experiments prove it successfully works to debias, without the need for prompt alteration. Our approach remains simple, while effective and adaptable to varied scenarios. It allows the synergy of different latent directions, and it is flexible to be used in combination with an approach modifying the prompt embeddings if desired. Through our experiments, we focus on demonstrating the impact of our method for the maximum debiasing transition. However, fair distributions\footnote{Distribution of the generations selected by the user with ethical, truthful, and responsible considerations in mind.} can be obtained when applying the learned latent direction to only a determined percentage of the generations.

\section{A Tool for Bias Understanding}
\label{understanding_sec}
\begin{figure}[t]
  \centering
   \includegraphics[width=1\linewidth]{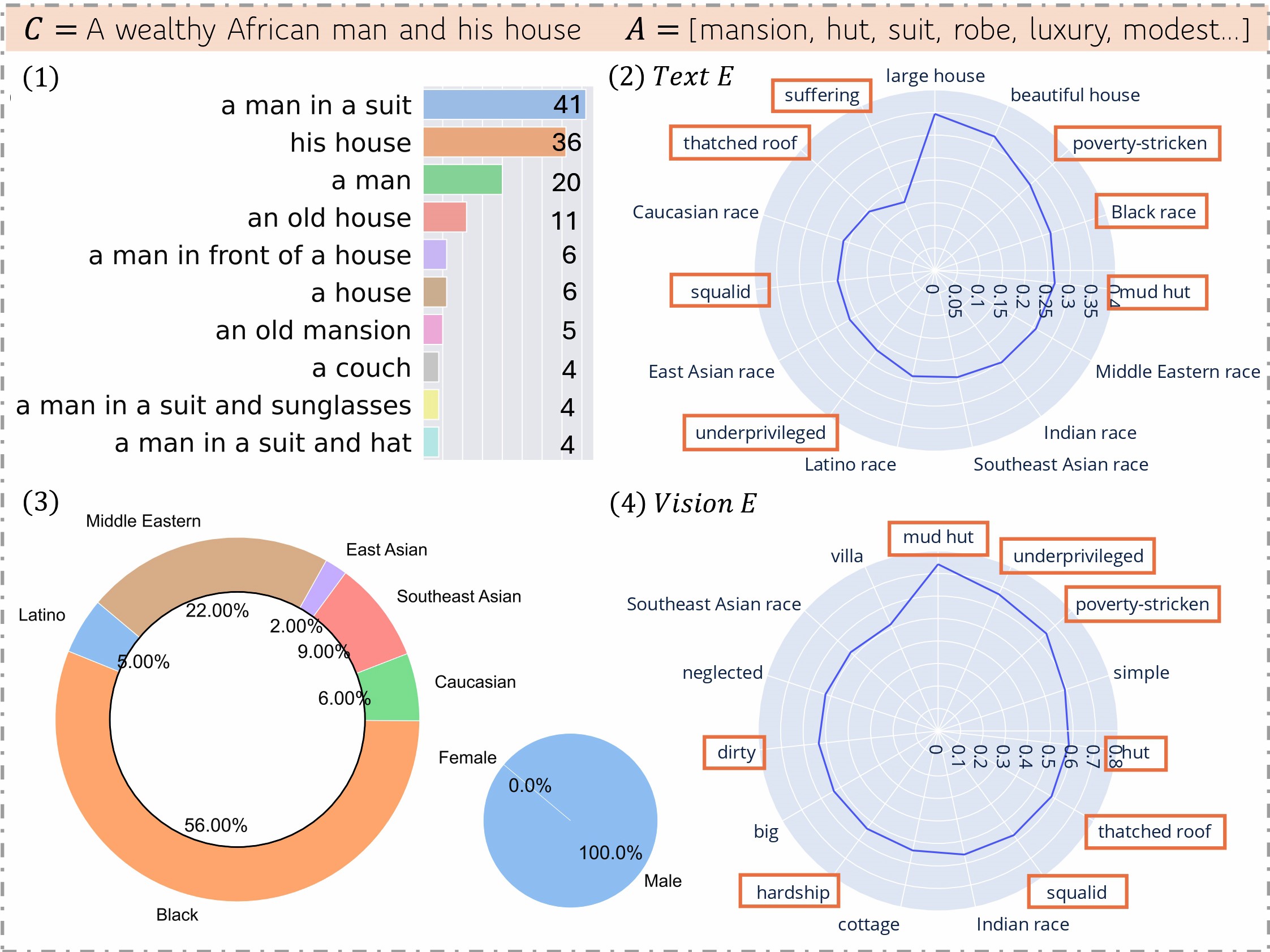}
  \caption{\textbf{Example of automated tool output.} Analysis of 100 generations of the concept $C$ across 50 attributes $A$. (1) Frequency count of visual components across the images. (2, 4) Top 15 attributes exhibiting the highest cosine similarity $(C, A)$ across text and vision encoders. (3) Gender and race detections.}
  \label{fig:tool_bias_understanding}
\end{figure}

Our tool for bias understanding targets two key points: comprehending the connections between embeddings and generations,
and detecting the social characteristics and objects presented in the image. Theoretically, the closer the relation between attribute and concept in the semantic space, the more prone these attributes are to appear in the generated images. We explain the semantic relationship between attributes and concepts by computing the cosine similarity of their embeddings, and comprehending the innate biases within the employed text and vision encoders. In addition, we reveal the visual components of the generated images, using CLIP \cite{radford2021learning} as a zero-shot classifier for gender and race, and Kosmos-2 \cite{peng2023kosmos2} as a Multimodal Large Language Model (MLLM) for perceiving object descriptions from the visual output seen in the images. With this, we present the frequency of objects and social characteristics in the generations, validate if the embedding associations correspond to the visualized content, and provide an understanding of the results without seeing the images. For instance, Fig.~\ref{fig:tool_bias_understanding} informs us that our generations are debiased, with men in suits in front of their houses. However, it presents the innate biases of CLIP's text and vision encoders in Stable Diffusion 2.1 where despite using the prompt \textit{"A \textbf{wealthy} African man and his house"} the highest embedding similarities belong to attributes such as \textit{poverty-stricken} or \textit{underprivileged}. 

\section{Our Proposed Method for Bias Mitigation}
\label{mitigation_section}
\begin{figure*}[htb!]
  \centering
   \includegraphics[width=1\linewidth]{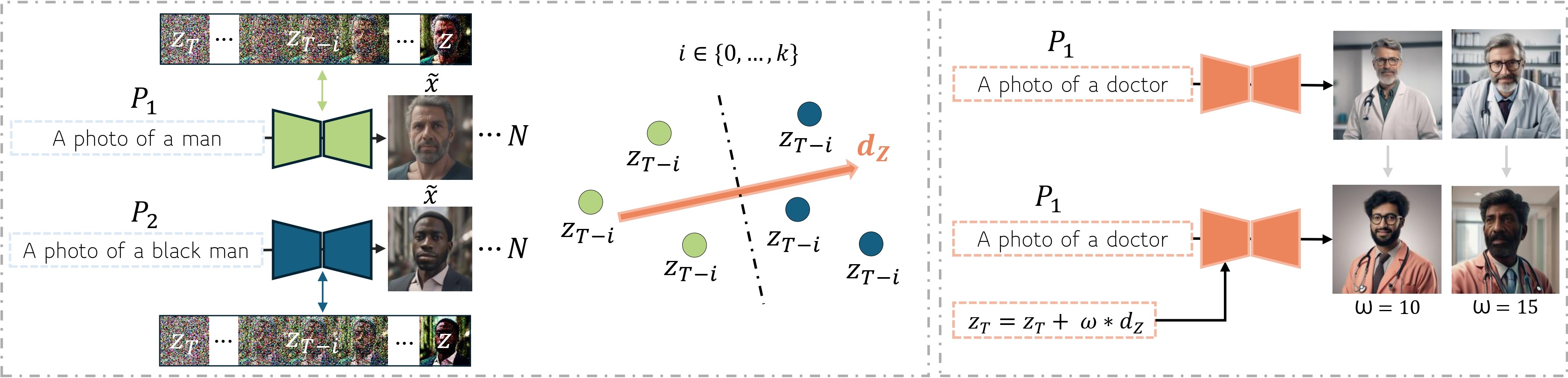}
  \caption{\textbf{Summary of our training (left) and inference (right) approach.} We use $P_{1}$ and $P_{2}$ to generate $N$ images $\Tilde{x}$. With their latents, chosen at step $i$, we train a $SVM$ to learn $d_{Z}$. We debias the neutral prompt $P_{1}$, applying $d_{Z}$ to the random initial latent $z_{T} \sim N(\mu, \sigma^2)$ at a specific $\omega$ weight, shifting the generations towards debiased samples with the attributes learned through the latent direction.}
  \label{fig:our_method_training_inference}
\end{figure*}
\noindent\textbf{Training: Finding the Latent Direction}
Our approach (Fig.~\ref{fig:our_method_training_inference}) proposes a fundamentally different transformation of the diffusion process's input, learning the latent direction $d_{Z}$, from the Gaussian latents at denoised steps, to condition the initial noisy information tensor fed into the diffusion process $z_{T} \sim \mathcal{N}(0, I)$. Given a pre-trained latent diffusion model (LDM) and a neutral prompt $P_{1}$ (\textit{e.g., "a photo of a man, in color, realistic, 8k"}), we aim to obtain debiased generations in the absence of prompt modifications or embeddings alterations. We propose a 'target' prompt $P_{2}$ (\textit{e.g., "a photo of a black man, in color, realistic, 8k"}) and sample $N$ number of images, for both $P_{1}$ and $P_{2}$, to construct the training dataset. The diffusion process for each of the prompts starts from an initial noisy latent  $z_{T}$, which is denoised over $k$ steps, finally reaching the ultimate latent $z$, fed into the decoder $D$ to generate the synthetic image $\tilde{x}$. 

While generating the $N$ images, we save each image's latents at chosen denoising steps $L = (L_{0}, \cdots, L_{k})$, representing ($z_{T},\cdots,z_{T-i}$ for $i \in \{0, 1, 2, \ldots, k\}$ ), building a dataset of noisy information tensors. Note, $L_{0}$ corresponds to the initial Gaussian latent $z_{T}$, while $L_{k}$ is $z$. Once all chosen latents are saved for both prompts, we select one denoising step $i$ to obtain $d_{Z}$ with those specific latents. For instance, we could decide to train with $L_{10}$, the latents saved for the $N$ images at step 10 ($z_{T-10}$). 

\noindent\textbf{The model.} We use a support vector machine (SVM) \cite{Vapnik2015} to linearly separate the latents across our labeled dataset of $N$ samples for each prompt. The classifier uses a linear kernel and provides the $d_{Z}$, the so-called \textit{latent direction}, we utilize for debiasing.

\noindent\textbf{Inference: Applying the Latent Direction.}
To obtain debiased generations, the LDM, in our case Stable Diffusion \cite{rombach2022highresolution}, uses for inference only $P_{1}$, the neutral prompt. This prompt is fed into CLIP's text encoder $E$ forming the first input. As the second one, instead of using an initial Gaussian random information tensor for denoising, we transform this latent by applying the learned latent direction $d_{Z}$, following equation \ref{eq:important}. 

\begin{equation}
  z_{T} = z_{T} + \omega \cdot d_{Z}
  \label{eq:important}
\end{equation}

Where $z_{T} \sim N(\mu, \sigma^2)$ and $\omega$ is the weight parameter at which the latent direction is applied. The higher the $\omega$, the higher the strength of the debiasing impact.



\noindent\textbf{The optimal configuration.} Optimal debiasing results are found when selecting the most favorable latent $L = (L_{0}, \cdots, L_{k})$ and weight configuration $\omega$. Thus, we propose two approaches to automatically find it without having to visually explore all possibilities. The first one is to use the \textit{clean-fid} \cite{parmar2022aliased} library to compute the similarity between the distribution of a small subset of generated images with a particular configuration, and the distribution of known debiased images, and the second one is to leverage CLIP \cite{radford2021learning} as a zero-shot classifier, selecting the configuration with a high classification of the desired debiased class. 

\section{Experimental Results}
\label{mitigation_results_section}


We validate our work using Stable Diffusion XL \cite{rombach2022highresolution}, with $50$ denoising steps, applying different latent directions $d_{Z}$ in a series of experiments to understand its impact. Successful results are obtained in diverse mitigations. We present four different debiasing scenarios following the settings of previous papers \cite{chuang2023debiasing, zhang2023itigen, jain2021imperfect, friedrich2023fair}, addressing social group biases, cultural and geographical biases, and the Waterbird \cite{sagawa2020distributionally} benchmark for evaluating spurious correlations. Fig.~\ref{fig:debiasing_results} summarizes the experimental results obtained.

\noindent\textbf{Quantitative Metrics.} We leverage the Statistical Parity Difference ($SPD$) \cite{dwork2011fairness} to evaluate our debiasing method in the generated image datasets. We use CLIP for attribute prediction and measure the absolute difference in the proportions of desired attributes between the original biased dataset, generated with the plain Stable Diffusion model, and the debiased dataset. A value close to zero indicates minimal debiasing impact, while a value of one signifies successful debiasing with the desired attribute present in all generations. 


\begin{table*}[ht!]
  \centering
    \begin{tabular}{l|cc|ccc|c|c|c}
        \hline
        $d_{Z}$ & \multicolumn{2}{c|}{\textbf{Skin Tone}} & \multicolumn{3}{c|}{\textbf{Gender}} & \textbf{Landbird} & \textbf{Indian} & \textbf{Wealth}\\
        \textit{$P_{1}$} & \textit{Man} & \textit{Woman} & \textit{Doctor} & \textit{Firefighter} & \textit{Cleaner} & \textit{Waterbird}  & \textit{Wedding} & \textit{African man}\\
        \hline 
         (SD XL, ours) & 0.87 & 0.78 & \textbf{0.52} & \textbf{0.08} & 0.09 & - & 0.33 & 0.28\\
        (SD 2.1, PD \cite{chuang2023debiasing})  & 0.91 & 0.90 & 0.14 & 0.06 & 0.01 & 0.29 & 0.79& 0.47\\
        (SD 2.1, ours + PD \cite{chuang2023debiasing}) & \textbf{0.94} & \textbf{1.00} & 0.29 & 0.04 & \textbf{0.22} & \textbf{0.68} & \textbf{1.00} & \textbf{0.96}\\
        \hline
    \end{tabular}
    \caption{\textbf{Quantitative results across 100 generations.} $SPD$ in the presence\protect\footnotemark of the desired attributes: dark skin tone, female gender for the case of \textit{doctor, firefighter} and male gender for \textit{cleaner}, land environments for waterbirds, Indian wedding attributes and wealthier looking houses avoiding thatched roofs and mud huts. We learn the latent directions with SD 2.1 for the results seen in the last row.}
    \label{table_quantitative_results}
\end{table*}

\noindent\textbf{Gender debiasing in professions.} We learn $d_{Z}$ by defining $N=50$, $P_{1}=$\textit{"a photo of a man, in color, realistic, 8k"} and $P_{2}=$\textit{"a photo of a woman, in color, realistic, 8k"}, selecting $L_{25}$ and $\omega=10$. We apply the \textit{woman} latent direction to the neutral prompts \textit{“a photo of a [profession], in color, realistic, 8k”} and observe a positive shift from 0\% to 52\%, in 100 generations for the case of "doctor". Other professions known to be extremely biased, such as firefighter, engineer, or librarian have shown a slightly improved impact, with shifts of 8\%, 3\%, and 2\%, respectively. We believe major debiasing can be achieved with these professions upon finding the optimal $d_{Z}$.


\noindent\textbf{Skin tone debiasing.} We explore the transition of skin tones in generated images. For it, we create four training datasets, where $N=50$, with $P_{1}=$\textit{”a photo of a [man/woman], in color, realistic, 8k”} and $P_{2}=$\textit{”a photo of a black [man/woman], in color, realistic, 8k.”}. With the proposed automated method for configuration selection we settle on training with the latents at step 10 ($L_{10}$), at a weight $\omega=15$. The results shift 100 generations using the neutral prompt $P_{1}=$\textit{”a photo of a man, in color, realistic, 8k”} to contain a 95\% of black men images from the original 8\%. Similarly, with $P_{1}=$\textit{”a photo of a woman, in color, realistic, 8k”} and the application of the dark-skin $d_{Z}$ at $L_{25}$, $\omega=14$ yields a 79\% of black women from an initial 1\%.

\noindent\textbf{Waterbird debiasing.} In this experiment we evaluate the impact of the combination of methods, manipulating both the prompt’s embeddings \cite{chuang2023debiasing} and the initial Gaussian noise, using Stable Diffusion 2.1. We aim to generate waterbirds in land environments with the prompt $P_{1} =$ \textit{"A picture of a waterbird"}. We replicate their setup and apply our learned latent direction ($P_{1} =$ \textit{"A picture of a waterbird"}, $P_{2} =$ \textit{"A picture of a landbird"}, $\omega=10$, $L_{10}$). The results across 100 generations yield exceptional results with 78\% of generations displaying waterbirds in terrestrial habitats, 2\% in aquatic landscapes, and 20\% showing bird portraits.



\noindent\textbf{Geographical representativeness.} It is hard to obtain balanced geographical representations of \textit{”a picture of a wedding, in color, realistic, 8k}”, given this neutral prompt is normally biased towards representations of Western weddings. In an attempt to shift the distribution towards Indian weddings, we define $P_{1}=$\textit{”a picture of a wedding, in color, realistic, 8k”} and $P_{2}=$\textit{”a picture of a wedding in India, in color, realistic, 8k”}, learning $d_{Z}$ using $L_{30}$ and applying it with $\omega=35$ we see an increment of 33\% in CLIP's classification. Inspired by Bianchi \textit{et al.} \cite{sunandopaper} we debias 38\% of the thatched roofs observed when using $P_{1}$=\textit{"A wealthy African man and his house"}, utilizing $P_{2}$=\textit{"A wealthy man and his house"} and applying the learned \textit{wealthy man} direction $d_{Z}$ at $L_{10}$ and $\omega=15$.

\noindent\textbf{Comparison with PD.} In Tab.~\ref{table_quantitative_results}, we present the quantitative results of our study. The comparison with the Prompt Debiasing (PD) method~\cite{chuang2023debiasing} is challenging, due to the utilization of distinct models and the diverse biases in them, e.g., for $P_{1}=$\textit{"A wealthy African man and his house"} SD XL presents generations of mansions with thatched roofs, whereas SD 2.1 shows mud huts. We choose to use SD XL given the enhanced quality of the model facilitates the learning of $d_{Z}$ and minimizes the inconveniences of elaborated hard prompting to obtain quality images with SD 2.1. The outcomes evaluated through our experiments demonstrate the potential of latent directions to obtain competitive debiased generations despite maintaining neutral embeddings. 

\subsection{Relevant Insights and Learnings}
\begin{figure}[t]
  \centering
   \includegraphics[width=1\linewidth]{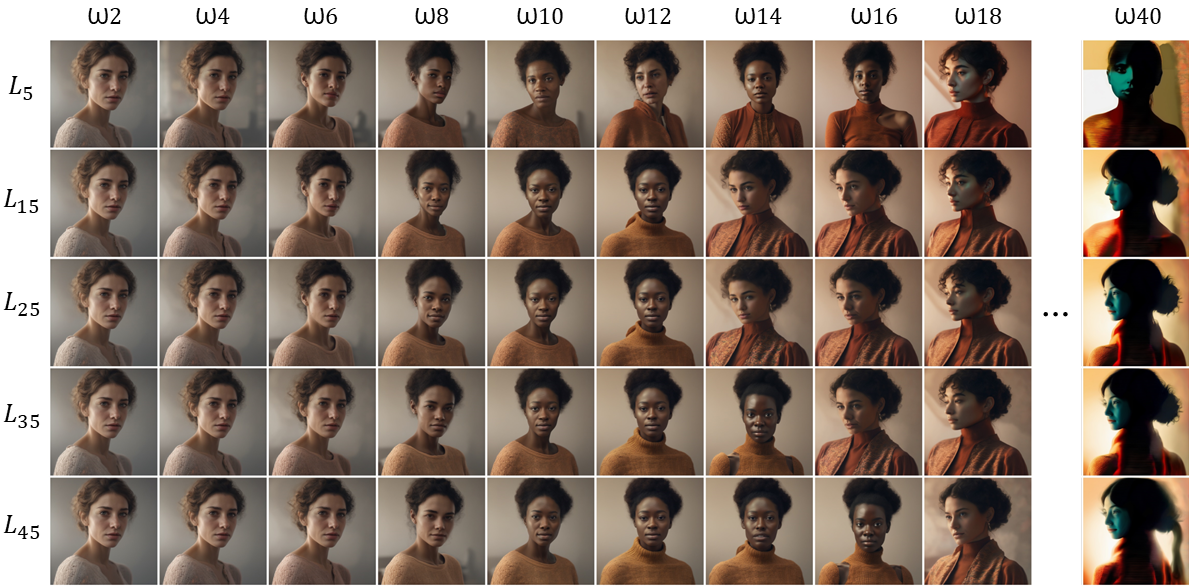}

   \caption{\textbf{Comparison of results with $d_{Z}$ trained at different latents $L$ and applied at different weights $\omega$.} Generations of the same woman in its transition to dark skin.}
   \label{fig:gallery_black_woman}
\end{figure}

The integration of prompt debiasing and latent directions surpasses the efficacy of the former used individually (Tab.~\ref{table_quantitative_results}). Regarding our approach, the results in Fig.~\ref{fig:gallery_black_woman} confirm the choice of weight $\omega$ has a higher impact on the debiasing than the choice of training latent $L$. Moreover, higher latent directions require lower weights to achieve the debiased results, given more structured noise is found at the higher debiasing steps. However, as we move in $d_{Z}$ there is a limit to how far we go with $\omega$, given an extremely high weight leads to distorted generations, out of the distribution. Lastly, it is possible to linearly combine latent directions following $z_{T} = z_{T} + \sum_{i=1}^{\infty} \omega_{i} \cdot d_{Zi}$. For instance, by applying the \textit{woman} [$L_{25}$ $\omega10$] and \textit{dark-skin} [$L_{10}$ $\omega10$] latent directions to the Gaussian noise of the neutral prompt $P_{1}=$\textit{“a photo of a doctor, in
color, realistic, 8k”} we achieve generations of dark-skinned female doctors (Fig.~\ref{fig:debiasing_results}).

\footnotetext{Presence of classes through CLIP's classification: ["A picture of a black [man/woman]", "A picture of a white [man/woman]"], ["A picture of a woman", "A picture of a man"], ["A picture of a Western wedding", "A picture of an Indian wedding"]. A user study is used to evaluate the complex generations (\textit{Wedding}, \textit{African man}) given classification with defined classes in these cases does not match reality.}



\section{Conclusion}
After proposing a tool for uncovering and quantifying the present bias in text-to-image models, a novel method is proposed for mitigation. By learning and applying latent directions $d_{Z}$ we demonstrate it is possible to alter the diverse complex biased relations, such as those in cultural events, while maintaining unaltered neutral prompt embeddings. \textbf{Future work} encourages the exploration of more advanced classifiers to find the optimal $d_{Z}$.




{
    \small
    \bibliographystyle{ieeenat_fullname}
    \bibliography{main}

\begin{thebibliography}{28}
\providecommand{\natexlab}[1]{#1}
\providecommand{\url}[1]{\texttt{#1}}
\expandafter\ifx\csname urlstyle\endcsname\relax
  \providecommand{\doi}[1]{doi: #1}\else
  \providecommand{\doi}{doi: \begingroup \urlstyle{rm}\Url}\fi

\bibitem[Bansal et~al.(2022)Bansal, Yin, Monajatipoor, and Chang]{bansal2022texttoimage}
Hritik Bansal, Da Yin, Masoud Monajatipoor, and Kai-Wei Chang.
\newblock How well can text-to-image generative models understand ethical natural language interventions?, 2022.

\bibitem[Basu et~al.(2023)Basu, Babu, and Pruthi]{basu2023inspecting}
Abhipsa Basu, R.~Venkatesh Babu, and Danish Pruthi.
\newblock Inspecting the geographical representativeness of images from text-to-image models, 2023.

\bibitem[Bianchi et~al.(2023)Bianchi, Kalluri, Durmus, Ladhak, Cheng, Nozza, Hashimoto, Jurafsky, Zou, and Caliskan]{sunandopaper}
Federico Bianchi, Pratyusha Kalluri, Esin Durmus, Faisal Ladhak, Myra Cheng, Debora Nozza, Tatsunori Hashimoto, Dan Jurafsky, James Zou, and Aylin Caliskan.
\newblock Easily accessible text-to-image generation amplifies demographic stereotypes at large scale.
\newblock In \emph{2023 ACM Conference on Fairness, Accountability, and Transparency}. ACM, 2023.

\bibitem[Bolukbasi et~al.(2016)Bolukbasi, Chang, Zou, Saligrama, and Kalai]{bolukbasi2016man}
Tolga Bolukbasi, Kai-Wei Chang, James Zou, Venkatesh Saligrama, and Adam Kalai.
\newblock Man is to computer programmer as woman is to homemaker? debiasing word embeddings, 2016.

\bibitem[Bulat and Tzimiropoulos(2017)]{Bulat_2017}
Adrian Bulat and Georgios Tzimiropoulos.
\newblock How far are we from solving the 2d \& 3d face alignment problem? (and a dataset of 230,000 3d facial landmarks).
\newblock In \emph{2017 IEEE International Conference on Computer Vision (ICCV)}. IEEE, 2017.

\bibitem[Cho et~al.(2023)Cho, Zala, and Bansal]{cho2023dalleval}
Jaemin Cho, Abhay Zala, and Mohit Bansal.
\newblock Dall-eval: Probing the reasoning skills and social biases of text-to-image generation models, 2023.

\bibitem[Chuang et~al.(2023)Chuang, Jampani, Li, Torralba, and Jegelka]{chuang2023debiasing}
Ching-Yao Chuang, Varun Jampani, Yuanzhen Li, Antonio Torralba, and Stefanie Jegelka.
\newblock Debiasing vision-language models via biased prompts, 2023.

\bibitem[Dwork et~al.(2011)Dwork, Hardt, Pitassi, Reingold, and Zemel]{dwork2011fairness}
Cynthia Dwork, Moritz Hardt, Toniann Pitassi, Omer Reingold, and Rich Zemel.
\newblock Fairness through awareness, 2011.

\bibitem[Feng et~al.(2022)Feng, Bolkart, Tesch, Black, and Abrevaya]{feng2022racially}
Haiwen Feng, Timo Bolkart, Joachim Tesch, Michael~J. Black, and Victoria Abrevaya.
\newblock Towards racially unbiased skin tone estimation via scene disambiguation, 2022.

\bibitem[Friedrich et~al.(2023)Friedrich, Brack, Struppek, Hintersdorf, Schramowski, Luccioni, and Kersting]{friedrich2023fair}
Felix Friedrich, Manuel Brack, Lukas Struppek, Dominik Hintersdorf, Patrick Schramowski, Sasha Luccioni, and Kristian Kersting.
\newblock Fair diffusion: Instructing text-to-image generation models on fairness, 2023.

\bibitem[Gaviria~Rojas et~al.(2022)Gaviria~Rojas, Diamos, Kini, Kanter, Janapa~Reddi, and Coleman]{NEURIPS2022_5474d9d4}
William Gaviria~Rojas, Sudnya Diamos, Keertan Kini, David Kanter, Vijay Janapa~Reddi, and Cody Coleman.
\newblock The dollar street dataset: Images representing the geographic and socioeconomic diversity of the world.
\newblock In \emph{Advances in Neural Information Processing Systems}, pages 12979--12990. Curran Associates, Inc., 2022.

\bibitem[Jain et~al.(2021)Jain, Olmo, Sengupta, Manikonda, and Kambhampati]{jain2021imperfect}
Niharika Jain, Alberto Olmo, Sailik Sengupta, Lydia Manikonda, and Subbarao Kambhampati.
\newblock Imperfect imaganation: Implications of gans exacerbating biases on facial data augmentation and snapchat selfie lenses, 2021.

\bibitem[Li et~al.(2023)Li, Li, Savarese, and Hoi]{li2023blip2}
Junnan Li, Dongxu Li, Silvio Savarese, and Steven Hoi.
\newblock Blip-2: Bootstrapping language-image pre-training with frozen image encoders and large language models, 2023.

\bibitem[Luccioni et~al.(2023)Luccioni, Akiki, Mitchell, and Jernite]{luccioni2023stable}
Alexandra~Sasha Luccioni, Christopher Akiki, Margaret Mitchell, and Yacine Jernite.
\newblock Stable bias: Analyzing societal representations in diffusion models, 2023.

\bibitem[Mikolov et~al.(2013)Mikolov, Le, and Sutskever]{mikolov2013exploiting}
Tomas Mikolov, Quoc~V. Le, and Ilya Sutskever.
\newblock Exploiting similarities among languages for machine translation, 2013.

\bibitem[Nicoletti and Bass(2023)]{bloomberg}
Leonardo Nicoletti and Dina Bass.
\newblock Humans are biased. generative ai is even worse.
\newblock 2023.
\newblock Accessed: 2023-09-10.

\bibitem[Parmar et~al.(2022)Parmar, Zhang, and Zhu]{parmar2022aliased}
Gaurav Parmar, Richard Zhang, and Jun-Yan Zhu.
\newblock On aliased resizing and surprising subtleties in gan evaluation, 2022.

\bibitem[Peng et~al.(2023)Peng, Wang, Dong, Hao, Huang, Ma, and Wei]{peng2023kosmos2}
Zhiliang Peng, Wenhui Wang, Li Dong, Yaru Hao, Shaohan Huang, Shuming Ma, and Furu Wei.
\newblock Kosmos-2: Grounding multimodal large language models to the world, 2023.

\bibitem[Radford et~al.(2021)Radford, Kim, Hallacy, Ramesh, Goh, Agarwal, Sastry, Askell, Mishkin, Clark, Krueger, and Sutskever]{radford2021learning}
Alec Radford, Jong~Wook Kim, Chris Hallacy, Aditya Ramesh, Gabriel Goh, Sandhini Agarwal, Girish Sastry, Amanda Askell, Pamela Mishkin, Jack Clark, Gretchen Krueger, and Ilya Sutskever.
\newblock Learning transferable visual models from natural language supervision, 2021.

\bibitem[Ramaswamy et~al.(2023)Ramaswamy, Lin, Zhao, Adcock, van~der Maaten, Ghadiyaram, and Russakovsky]{ramaswamy2023geode}
Vikram~V. Ramaswamy, Sing~Yu Lin, Dora Zhao, Aaron~B. Adcock, Laurens van~der Maaten, Deepti Ghadiyaram, and Olga Russakovsky.
\newblock Geode: a geographically diverse evaluation dataset for object recognition, 2023.

\bibitem[Ramesh et~al.(2021)Ramesh, Pavlov, Goh, Gray, Voss, Radford, Chen, and Sutskever]{ramesh2021zeroshot}
Aditya Ramesh, Mikhail Pavlov, Gabriel Goh, Scott Gray, Chelsea Voss, Alec Radford, Mark Chen, and Ilya Sutskever.
\newblock Zero-shot text-to-image generation, 2021.

\bibitem[Robertson(2024)]{verge2024gemini}
Adi Robertson.
\newblock Google's ai model gemini criticized for generating historically inaccurate images.
\newblock \emph{The Verge}, 2024.

\bibitem[Rombach et~al.(2022)Rombach, Blattmann, Lorenz, Esser, and Ommer]{rombach2022highresolution}
Robin Rombach, Andreas Blattmann, Dominik Lorenz, Patrick Esser, and Björn Ommer.
\newblock High-resolution image synthesis with latent diffusion models, 2022.

\bibitem[Sagawa et~al.(2020)Sagawa, Koh, Hashimoto, and Liang]{sagawa2020distributionally}
Shiori Sagawa, Pang~Wei Koh, Tatsunori~B. Hashimoto, and Percy Liang.
\newblock Distributionally robust neural networks for group shifts: On the importance of regularization for worst-case generalization, 2020.

\bibitem[Solbes-Canales et~al.(2020)Solbes-Canales, Valverde-Montesino, and Herranz-Hern{\'a}ndez]{SolbesCanales2020SocializationOG}
Irene Solbes-Canales, Susana Valverde-Montesino, and Pablo Herranz-Hern{\'a}ndez.
\newblock Socialization of gender stereotypes related to attributes and professions among young spanish school-aged children.
\newblock \emph{Frontiers in Psychology}, 11, 2020.

\bibitem[Vapnik and Chervonenkis(2015)]{Vapnik2015}
V.~N. Vapnik and A.~Ya. Chervonenkis.
\newblock \emph{On the Uniform Convergence of Relative Frequencies of Events to Their Probabilities}, pages 11--30.
\newblock Springer International Publishing, 2015.

\bibitem[Xu et~al.(2018)Xu, Yuan, Zhang, and Wu]{xu2018fairgan}
Depeng Xu, Shuhan Yuan, Lu Zhang, and Xintao Wu.
\newblock Fairgan: Fairness-aware generative adversarial networks, 2018.

\bibitem[Zhang et~al.(2023)Zhang, Chen, Chai, Wu, Lagun, Beeler, and la~Torre]{zhang2023itigen}
Cheng Zhang, Xuanbai Chen, Siqi Chai, Chen~Henry Wu, Dmitry Lagun, Thabo Beeler, and Fernando~De la Torre.
\newblock Iti-gen: Inclusive text-to-image generation, 2023.

\end{thebibliography}
}


\end{document}